# A General Framework for Learning Algebraic Properties from Cayley Graphs using Graph Neural Networks

Tal Weissblat

Email: tal.weisblat@gmail.com

## Abstract

In this work, we present a general Graph Neural Network (GNN) framework for learning algebraic properties of finite groups from their Cayley graph representations. The framework provides a unified computational pipeline consisting of a common graph construction procedure, feature representation, training methodology, and GNN architecture, with only the target labeling function varying across classification tasks. To demonstrate the generality of the proposed approach, we consider three representative algebraic properties: abelianity, nilpotency, and solvability.

Experiments were conducted on a benchmark of 176 finite groups drawn from several classical families, with all groups included in each classification task. To address class imbalance, class-weighted cross-entropy loss was employed where appropriate during training. Furthermore, the family $\mathrm{PSL}(2, \mathrm{q})$ was reserved exclusively for testing, enabling evaluation of the framework's ability to generalize to previously unseen group families.

The best-performing models achieved test balanced accuracies of 1.000, 0.856, and 0.875 for abelianity, nilpotency, and solvability, respectively. Although the same computational framework was employed across all tasks, different GNN architectures proved optimal for different algebraic properties, suggesting that the representational complexity required to learn a property depends on its underlying algebraic structure. These results demonstrate that GNNs can effectively learn multiple algebraic properties directly from Cayley graph representations while exhibiting strong generalization to unseen group families. More broadly, the proposed framework establishes a computational methodology for studying algebraic properties of finite groups using graph-based machine learning.

# 1 Introduction

Graph neural networks (GNNs) have emerged as a powerful class of machine learning (ML) models for learning from graph-structured data [1,2]. By exploiting the relational structure of graphs, GNNs have achieved remarkable success in applications such as social-network analysis, molecular property prediction, recommendation systems, and scientific computing. More broadly, these advances demonstrate the potential of graph representation learning to extract meaningful structural information from complex discrete objects.

Graphs also play a fundamental role in algebra and group theory, where they provide natural representations of algebraic structures. One of the most important examples is the Cayley graph, which represents a group with respect to a generating set and encodes important aspects of its algebraic structure through its topology. Cayley graphs have been extensively studied in group theory, combinatorics, and computer science, making them a natural setting for applying graph-based learning methods to algebraic problems [3,4].

In previous work [5], we introduced a GNN framework for predicting the solvability of finite groups from their Cayley graph representations. Using only structural information extracted from Cayley graphs, the model successfully distinguished solvable from non-solvable groups, demonstrating that GNNs can learn nontrivial algebraic properties directly from graph representations.

However, solvability is only one of several fundamental properties that characterize the structure of finite groups. Other important properties, including abelianity and nilpotency, occupy central positions in the classical hierarchy of finite-group theory [6]. While our previous work showed that solvability can be learned from Cayley graphs, it remains unclear whether the same methodology can be generalized into a unified framework for learning multiple algebraic properties. Addressing this question is important for understanding whether graph representation learning captures property-specific patterns or more general structural information about finite groups.

In this paper, we address this problem by developing a property-independent GNN framework for learning algebraic properties of finite groups from their Cayley graph representations. Rather than designing separate models for individual properties, we employ a common graph construction procedure, feature representation, model architecture, and training pipeline, varying only the target labeling function. The framework is demonstrated on three representative properties—abelianity, nilpotency, and solvability—and its ability to generalize is evaluated on the previously unseen family $\mathrm{PSL}(2, \mathrm{q})$, which is excluded entirely from training and validation.

Compared with our previous work [5], the experimental benchmark is substantially expanded through the inclusion of additional finite-group families and numerous direct-product constructions. The resulting benchmark comprises 176 finite groups, all of which participate in each classification task. Furthermore, the family $\mathrm{PSL}(2, \mathrm{q})$ is reserved exclusively for testing, providing a stronger assessment of the framework's ability to generalize to previously unseen algebraic structures.

The main contributions of this paper are fourfold. First, we generalize the solvability-specific framework introduced in [5] into a property-independent methodology for learning multiple algebraic properties from Cayley graphs. Second, we demonstrate the framework on three fundamental properties of finite groups—abelianity, nilpotency, and solvability—using a common graph construction procedure, feature representation, and training pipeline. Third, we evaluate the framework on an expanded benchmark comprising 176 finite groups, including additional group families and direct-product constructions. Fourth, we demonstrate that the learned models generalize to the previously unseen family $\mathrm{PSL}(2, \mathrm{q})$, which is excluded entirely from training and validation.

By applying a unified graph-learning methodology across multiple algebraic properties and evaluating its performance on a previously unseen family of finite groups, this work investigates the extent to which algebraic structure can be recovered from Cayley graph representations. More broadly, it contributes to the intersection of group theory, graph theory, and machine learning, providing a foundation for future research on learning algebraic structures from graph-based representations.

## 2 Group-Theoretic Background

In this section, we introduce the mathematical structures and algebraic properties used throughout this work. We first define Cayley graphs, which provide graph representations of finite groups and serve as inputs to our learning framework. We then define the three group-theoretic properties considered in this work—abelianity, nilpotency, and solvability—which serve as the target labels for the classification tasks [6].

**Cayley graph.** Let $G$ be a finite group generated by a set $S$. The Cayley graph $\mathrm{Cay}(G, S \cup S^{-1})$ is the undirected graph whose vertices are the elements of $G$. Two vertices $g, h \in G$ are adjacent if $h = gs$ for some $s \in S \cup S^{-1}$.

**Abelian group.** A group $G$ is abelian if $ab = ba$ for all $a, b \in G$, that is, every pair of elements commutes.

**Nilpotent group.** A group $G$ is nilpotent if its lower central series, defined by $\gamma_1(G) = G$ and $\gamma_{i+1}(G) = [G, \gamma_i(G)]$ for every $i \geq 1$, reaches the trivial group. Here, $[G, \gamma_i(G)]$ denotes the commutator group generated by all commutators $[a, b]$, where $a \in G$ and $b \in \gamma_i(G)$. Thus, $G$ is nilpotent if there exists $c > 0$ for which $\gamma_{c+1}(G) = \{e\}$.

**Solvable group.** A group $G$ is solvable if its derived series, defined by $G^{(0)} = G$ and $G^{(i+1)} =$

$[G^{(i)}, G^{(i)}]$ for every $i \geq 0$, reaches the trivial group. Thus, $G$ is solvable if there exists an integer $n \geq 0$ such that $G^{(n)} = \{e\}$.

These properties form a classical hierarchy in group theory: abelian $\Longrightarrow$ nilpotent $\Longrightarrow$ solvable Consequently, every abelian group is nilpotent, and every nilpotent group is solvable, while the converse implications do not generally hold.

## 3 Methodology

Building upon the solvability-prediction framework introduced in [5], we generalize this approach into a property-independent framework for predicting algebraic properties of finite groups from their Cayley graph representations. The graph construction procedure, feature representation, and training methodology remain unchanged across all experiments, with only the target labels varying according to the property under consideration. Candidate GNN architectures are selected based on their validation performance from a common pool of architectures. An overview of the proposed framework is shown in Figure 1.

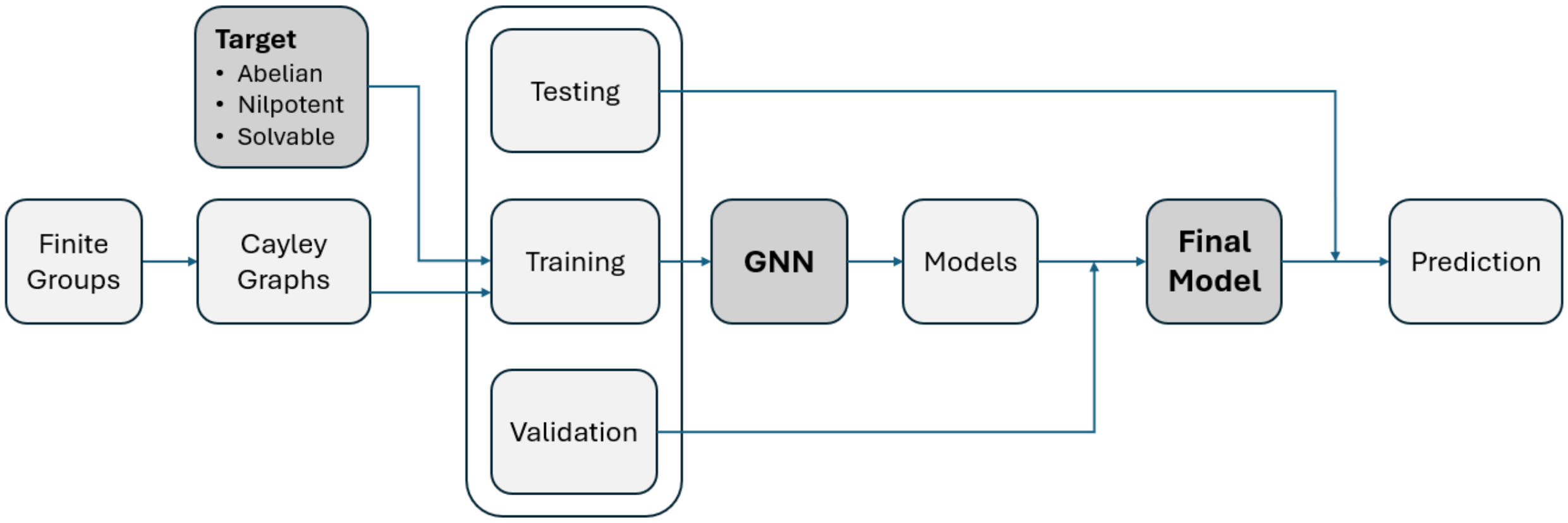

**Figure 1.** Overview of the proposed framework. Finite groups are converted into Cayley graphs, partitioned into training, validation, and test sets, and used to train GNN models. The target labels specify the algebraic property considered in the current classification task (abelianity, nilpotency, or solvability). The selected model is then used for prediction.

Each group is treated as an independent data instance and represented by its Cayley graph, which serves as input to a GNN classifier. For clarity, the methodology is organized into the following four stages:

1. Group generation
2. Cayley graph construction
3. Target labeling
4. GNN modeling

These stages are described in the following subsections.

## 3.1 Group Generation

The benchmark was constructed to include a broad range of algebraic structures while remaining computationally manageable. It comprises representatives from several classical families of finite groups together with direct-product constructions. The following summarizes the composition of the benchmark.

**Table 1.** Composition of the benchmark by group family.

| Group Family | Notation | Number of Groups |
|---|---|---|
| Cyclic | $C_n$ | 39 |
| Dihedral | $D_n$ | 23 |
| Quaternion | $Q_{2^m}$ | 5 |
| Symmetric | $S_n$ | 5 |
| Alternating | $A_n$ | 5 |
| Projective Special Linear | $PSL(2, q)$ | 7 |
| Direct Products | $G \times H$ | 92 |
| **Total** | | 176 |

The direct-product family consists of products of groups selected from the cyclic, dihedral, quaternion, symmetric, and alternating families. These groups exhibit different combinations of the algebraic properties considered in this work, namely abelianity, nilpotency, and solvability. Because the distribution of these properties varies across the benchmark, separate datasets were constructed for the three classification tasks while ensuring that each task includes representatives from every group family.

The inclusion of multiple group families and direct-product constructions provides a diverse collection of algebraic structures, reducing the likelihood that the learning process relies on family-specific characteristics rather than structural features associated with the target property. Compared with the benchmark in [5], the present benchmark substantially expands the collection of groups and extends the framework from a single property (solvability) to multiple algebraic properties.

## 3.2 Cayley Graph Construction

For each group family, a fixed symmetric generating set was used to construct the corresponding undirected Cayley graph. Cyclic groups were generated by a single generator, dihedral groups by a rotation and a reflection, symmetric groups by a transposition and an $n$-cycle, while quaternion, alternating, and projective special linear groups were constructed using their standard generating sets. For direct products $G \times H$, the generating set consisted of $(g, e)$ and $(e, h)$, where $g$ and $h$ belong to the generating sets of $G$ and $H$, respectively.

## 3.3 Property Labeling

The graph construction process is independent of the algebraic property being studied. Once the Cayley graphs have been generated, each graph is assigned a binary label corresponding to the target property. In this work, the proposed framework is evaluated on the following three binary classification tasks:

- Abelian versus non-abelian groups.
- Nilpotent versus non-nilpotent groups.
- Solvable versus non-solvable groups.

## 3.4 GNN Modeling

Once the Cayley graphs have been constructed and labeled, GNNs are used to predict the target algebraic property. This stage consists of three components: the GNN architecture, the dataset partitioning strategy, and the training and evaluation procedures. The same modeling framework is applied across all classification tasks, ensuring that differences in predictive performance across the three classification tasks arise from the target property rather than from changes in the learning methodology.

### 3.4.1 GNN Architecture

To evaluate the proposed framework, we employ a family of GNN architectures based on Graph Convolutional Network (GCN) layers. Each model receives a Cayley graph as input and produces a graph-level prediction of the target algebraic property. The network consists of two GCNConv layers with GELU activation functions. The resulting node representations are aggregated using global mean pooling to obtain a graph-level embedding, which is then passed through a fully connected hidden layer and a final output layer with two neurons corresponding to the classes of the binary classification task. To investigate the effect of network capacity, multiple GNN architectures with varying hidden dimensions were evaluated. The candidate architectures evaluated for each classification task are presented in the Results section.

### 3.4.2 Dataset Partitioning Strategy

Because the class distributions differ across the algebraic properties considered in this work, namely abelianity, nilpotency, and solvability, the exact composition of the training, validation, and test sets varies between classification tasks. Nevertheless, all 176 groups in the benchmark were included in each of the three classification tasks, and all dataset partitions were constructed according to the same general principles. Each dataset was partitioned into approximately 60% training, 20% validation, and 20% test samples. The validation set was used for model selection, while the test set retained its natural class distribution to assess predictive performance.

In each classification task, the $\mathrm{PSL}(2, \mathrm{q})$ family was excluded from the training and validation sets and reserved exclusively for testing. This design evaluates the ability of the GNN to generalize to a previously unseen family of finite groups.

### 3.4.3 Training Configuration

All models were trained using the Adam optimizer with a learning rate of 0.002 and the cross-entropy loss function. To address class imbalance, class-weighted cross-entropy loss was employed during training, with class weights determined from the class frequencies in the training set. Each model was trained for 100 epochs.

Preliminary experiments indicated that the distance-based node feature representation consistently outperformed the order-based representation. Consequently, all experiments reported in this paper use the distance-based representation. For each classification task, multiple GNN architectures were evaluated, and the architecture achieving the highest balanced accuracy (BA) on the validation set was selected for final evaluation on the test set. Apart from the target labels and the selected architecture, the same training procedure and hyperparameter configuration were used throughout.

### 3.4.4 Evaluation Metric

Model performance was evaluated using BA [9]. Let $\mathrm{TP}$, $\mathrm{TN}$, $\mathrm{FP}$, and $\mathrm{FN}$ denote the numbers of true positives, true negatives, false positives, and false negatives, respectively. BA is defined as

$$\mathrm{BA} = \frac{1}{2}\left(\frac{\mathrm{TP}}{\mathrm{TP} + \mathrm{FN}} + \frac{\mathrm{TN}}{\mathrm{TN} + \mathrm{FP}}\right)$$

BA was selected because the class distributions associated with the considered algebraic properties are not necessarily balanced. By assigning equal weight to both classes, it provides a more informative evaluation than standard accuracy for imbalanced classification tasks. For each classification task, BA was computed on the validation set for model selection and on the test set for the final evaluation.

## 4 Results

This section presents the experimental results obtained for the three binary classification tasks considered in this work: abelianity, nilpotency, and solvability. For each task, multiple GNN architectures were evaluated, and the architecture achieving the highest BA on the validation set was selected for final evaluation on the test set. The following summarizes the best-performing architecture and its corresponding predictive performance for each classification task.

**Table 2.** Summary of the best-performing GNN architectures across the three classification tasks.

| Property | Dataset (Pos./Neg.) | Train | Validation | Test | Best Architecture | Validation BA | Test BA |
|---|---|---|---|---|---|---|---|
| Abelian | 63/113 | 102 | 34 | 40 | (2-16-16) | 1.000 | 1.000 |
| Nilpotent | 95/81 | 101 | 34 | 41 | (2-64-64) | 0.768 | 0.856 |
| Solvable | 141/35 | 102 | 34 | 40 | (2-8-8) | 0.917 | 0.875 |

Overall, the proposed framework achieved strong predictive performance across all three classification tasks. Although the optimal GNN architecture differed between properties, each task admitted at least one architecture with high predictive performance. In particular, the framework achieved perfect classification for abelianity and strong predictive performance for both nilpotency and solvability, demonstrating that Cayley graph representations contain sufficient structural information to support accurate prediction of multiple algebraic properties. The following subsections examine each classification task in detail and analyze the corresponding model architectures and predictive performance.

## 4.1 Abelianity

The abelianity classification task is first evaluated by summarizing the dataset partition used in the experiments. Table 3 presents the distribution of abelian and non-abelian groups across the training, validation, and test sets.

**Table 3.** Distribution of the 176 benchmark groups across the training, validation, and test sets for the abelianity classification task.

| Set | Abelian | Non-Abelian | Total | Percentage |
|---|---|---|---|---|
| Training | 38 | 64 | 102 | 58% |
| Validation | 13 | 21 | 34 | 19.3% |
| Test | 12 | 28 | 40 | 22.7% |
| **Total** | 63 | 113 | 176 | 100% |

The following analysis compares the candidate GNN architectures and evaluates their predictive performance on the abelianity classification task.

**Table 4.** Validation performance of the candidate GNN architectures for the abelianity classification task.

| Arch. | TP | FN | TN | FP | BA |
|---|---|---|---|---|---|
| (2-8-8) | 12 | 1 | 19 | 2 | 0.914 |
| (2-16-16) | 13 | 0 | 21 | 0 | 1.000 |
| (2-32-32) | 13 | 0 | 21 | 0 | 1.000 |

| Arch. | TP | FN | TN | FP | BA |
|---|---|---|---|---|---|
| (2-64-32) | 13 | 0 | 21 | 0 | 1.000 |
| (2-64-64) | 13 | 0 | 21 | 0 | 1.000 |
| (2-128-128) | 13 | 0 | 21 | 0 | 1.000 |

Several candidate GNN architectures achieved a BA of 1.000 on the validation set. Among these, the architecture (2-16-16) was selected for evaluation on the test set because it is the simplest architecture achieving the highest validation performance.

**Table 5.** Test-set predictions produced by the selected GNN architecture (2-16-16).

| Group | True | Pred | Score |
|---|---|---|---|
| C3xC9 | 1 | 1 | 0.996 |
| C2xC6 | 1 | 1 | 1.000 |
| C8 | 1 | 1 | 1.000 |
| C4xC7 | 1 | 1 | 0.999 |
| C10 | 1 | 1 | 1.000 |

| Group | True | Pred | Score |
|---|---|---|---|
| D2xS7 | 0 | 0 | 0.000 |
| C4xD9 | 0 | 0 | 0.000 |
| C2xD4 | 0 | 0 | 0.000 |
| C3xD7 | 0 | 0 | 0.000 |
| A6 | 0 | 0 | 0.000 |

| Group | True | Pred | Score |
|---|---|---|---|
| C2xQ16 | 0 | 0 | 0.000 |
| D16 | 0 | 0 | 0.000 |
| D2xS5 | 0 | 0 | 0.000 |
| D3xQ32 | 0 | 0 | 0.000 |
| S7 | 0 | 0 | 0.000 |

| | | | | | | | | | | | |
|---|---|---|---|---|---|---|---|---|---|---|---|
| C16 | 1 | 1 | 1.000 | C3xQ32 | 0 | 0 | 0.000 | A8 | 0 | 0 | 0.000 |
| C17 | 1 | 1 | 1.000 | C3xD5 | 0 | 0 | 0.000 | PSL2_5 | 0 | 0 | 0.000 |
| C19 | 1 | 1 | 1.000 | C3xQ16 | 0 | 0 | 0.000 | PSL2_7 | 0 | 0 | 0.000 |
| C2xC10 | 1 | 1 | 0.996 | C2xQ8 | 0 | 0 | 0.000 | PSL2_11 | 0 | 0 | 0.000 |
| C3 | 1 | 1 | 1.000 | D10 | 0 | 0 | 0.010 | PSL2_13 | 0 | 0 | 0.000 |
| C9 | 1 | 1 | 1.000 | D2xA6 | 0 | 0 | 0.000 | PSL2_17 | 0 | 0 | 0.000 |
| C2xC3 | 1 | 1 | 0.746 | D2xQ32 | 0 | 0 | 0.000 | PSL2_19 | 0 | 0 | 0.000 |
| C4xD6 | 0 | 0 | 0.000 | C3xA8 | 0 | 0 | 0.000 | PSL2_23 | 0 | 0 | 0.000 |
| C4xQ8 | 0 | 0 | 0.000 | | | | | | | | |

The Score column reports the predicted probability assigned to the predicted class and therefore provides a measure of the model's confidence in each prediction. The model assigned high confidence to almost all predictions, including the previously unseen $\mathrm{PSL}(2, \mathrm{q})$ groups.

**Table 6.** Confusion matrix and BA of the selected GNN architecture (2-16-16) on the test set.

| Architecture | Abelian | | Non-abelian | | BA |
|---|---|---|---|---|---|
| | TP | FN | TN | FP | |
| (2-16-16) | 12 | 0 | 28 | 0 | 1.000 |

The selected architecture (2-16-16) correctly classified all 40 groups in the test set, achieving a BA of 1.000. These results demonstrate that the proposed framework accurately distinguishes between abelian and non-abelian groups and successfully generalizes to the previously unseen $\mathrm{PSL}(2, \mathrm{q})$ family.

## 4.2 Nilpotency

The nilpotency classification task is next evaluated by summarizing the dataset partition used in the experiments. The distribution of nilpotent and non-nilpotent groups across the training, validation, and test sets is shown below.

**Table 7.** Distribution of the 176 benchmark groups across the training, validation, and test sets for the nilpotency classification task.

| Set | Nilpotent | Non-Nilpotent | Total | Percentage |
|---|---|---|---|---|
| Training | 57 | 44 | 101 | 57.4% |
| Validation | 19 | 15 | 34 | 19.3% |
| Test | 19 | 22 | 41 | 23.3% |
| **Total** | 95 | 81 | 176 | 100% |

The following analysis compares the candidate GNN architectures and evaluates their predictive performance on the nilpotency classification task.

**Table 8.** Validation performance of the candidate GNN architectures

| Architecture | TP | FN | TN | FP | BA | Architecture | TP | FN | TN | FP | BA |
|---|---|---|---|---|---|---|---|---|---|---|---|
| (2-8-8) | 13 | 6 | 10 | 5 | 0.675 | (2-64-32) | 14 | 5 | 10 | 5 | 0.702 |
| (2-16-16) | 13 | 6 | 10 | 5 | 0.675 | (2-64-64) | 14 | 5 | 12 | 3 | 0.768 |
| (2-32-32) | 14 | 5 | 9 | 6 | 0.668 | (2-128-128) | 14 | 5 | 10 | 5 | 0.702 |

The architecture (2-64-64) achieved the highest validation BA and was therefore selected for evaluation on the test set.

**Table 9.** Test-set predictions produced by the selected GNN architecture (2-64-64).

| Group | True | Pred | Score | Group | True | Pred | Score | Group | True | Pred | Score |
|---|---|---|---|---|---|---|---|---|---|---|---|
| C4xC4 | 1 | 1 | 0.993 | C33 | 1 | 1 | 1.000 | C2xD5 | 0 | 0 | 0.374 |
| C31 | 1 | 1 | 1.000 | C37 | 1 | 1 | 1.000 | D3xQ8 | 0 | 1 | 0.598 |
| C29 | 1 | 1 | 1.000 | C5 | 1 | 1 | 1.000 | D19 | 0 | 0 | 0.062 |
| C4xQ16 | 1 | 1 | 0.568 | C16 | 1 | 1 | 1.000 | C3xA5 | 0 | 0 | 0.478 |
| D4xQ16 | 1 | 1 | 0.604 | C3xQ32 | 1 | 1 | 0.508 | A8 | 0 | 0 | 0.000 |
| C6 | 1 | 1 | 1.000 | C3xA4 | 0 | 1 | 0.541 | PSL2_5 | 0 | 0 | 0.358 |
| C2xC9 | 1 | 1 | 1.000 | D3xQ16 | 0 | 1 | 0.602 | PSL2_7 | 0 | 0 | 0.191 |
| C3xD4 | 1 | 0 | 0.394 | C4xD5 | 0 | 0 | 0.424 | PSL2_11 | 0 | 0 | 0.077 |
| C13 | 1 | 1 | 1.000 | C3xA8 | 0 | 0 | 0.001 | PSL2_13 | 0 | 0 | 0.036 |
| C4xC9 | 1 | 1 | 1.000 | D2xA4 | 0 | 1 | 0.556 | PSL2_17 | 0 | 0 | 0.010 |
| C4xQ64 | 1 | 0 | 0.186 | C4xD6 | 0 | 0 | 0.441 | PSL2_19 | 0 | 0 | 0.004 |
| C15 | 1 | 1 | 1.000 | A6 | 0 | 0 | 0.126 | PSL2_23 | 0 | 0 | 0.001 |
| C19 | 1 | 1 | 1.000 | D2xS3 | 0 | 0 | 0.394 | D12 | 0 | 0 | 0.180 |
| C30 | 1 | 1 | 1.000 | C4xD10 | 0 | 0 | 0.307 | | | | |

The Score column reports the predicted probability assigned to the predicted class and therefore provides a measure of the model's confidence in each prediction. Despite several misclassifications, the model correctly classified all previously unseen $\mathrm{PSL}(2, q)$ groups.

**Table 10.** Confusion matrix and BA of the selected GNN architecture (2-64-64) on the test set.

| Architecture | Nilpotent | | Non-Nilpotent | | BA |
|---|---|---|---|---|---|
| | TP | FN | TN | FP | |
| (2-64-64) | 17 | 2 | 18 | 4 | 0.856 |

The selected architecture (2-64-64) achieved a BA of 0.856 on the test set. These results demonstrate that the proposed framework can effectively distinguish between nilpotent and non-nilpotent groups while successfully generalizing to the previously unseen $\mathrm{PSL}(2, q)$ family.

## 4.3 Solvability

The solvability classification task is finally evaluated by summarizing the dataset partition used in the experiments. The distribution of solvable and non-solvable groups across the training, validation, and test sets is shown below.

**Table 11.** Distribution of the 176 benchmark groups across the training, validation, and test sets for the solvability classification task.

| Set | Solvable | Non-Solvable | Total | Percentage |
|---|---|---|---|---|
| Training | 85 | 17 | 102 | 58% |
| Validation | 28 | 6 | 34 | 19.3% |
| Test | 28 | 12 | 40 | 22.7% |
| **Total** | 141 | 35 | 176 | 100% |

The following analysis compares the candidate GNN architectures and evaluates their predictive performance on the solvability classification task.

**Table 12.** Validation performance of the candidate GNN architectures.

| Arch. | TP | FN | TN | FP | BA |
|---|---|---|---|---|---|
| (2-8-8) | 28 | 0 | 5 | 1 | 0.917 |
| (2-16-16) | 28 | 0 | 5 | 1 | 0.917 |
| (2-32-32) | 28 | 0 | 5 | 1 | 0.917 |

| Arch. | TP | FN | TN | FP | BA |
|---|---|---|---|---|---|
| (2-64-32) | 28 | 0 | 5 | 1 | 0.917 |
| (2-64-64) | 28 | 0 | 5 | 1 | 0.917 |
| (2-128-128) | 28 | 0 | 5 | 1 | 0.917 |

All candidate GNN architectures achieved identical validation performance, with a BA of 0.917. Accordingly, the simplest architecture, (2-8-8), was selected for evaluation on the test set.

**Table 13.** Test-set predictions produced by the selected GNN architecture (2-8-8).

| Group | True | Pred | Score |
|---|---|---|---|
| D4xQ16 | 1 | 1 | 0.974 |
| C2xC8 | 1 | 1 | 1.000 |
| D4xQ32 | 1 | 1 | 0.942 |
| D2xA4 | 1 | 1 | 0.967 |
| D16 | 1 | 1 | 1.000 |
| C2xC2 | 1 | 1 | 0.980 |
| C4xC7 | 1 | 1 | 1.000 |
| C3xC8 | 1 | 1 | 1.000 |
| C4xC9 | 1 | 1 | 1.000 |
| C26 | 1 | 1 | 1.000 |
| C2xC4 | 1 | 1 | 0.995 |
| C4 | 1 | 1 | 0.988 |
| C2xC10 | 1 | 1 | 1.000 |
| Q8 | 1 | 1 | 0.998 |

| Group | True | Pred | Score |
|---|---|---|---|
| D22 | 1 | 1 | 0.999 |
| D18 | 1 | 1 | 1.000 |
| C24 | 1 | 1 | 1.000 |
| C8 | 1 | 1 | 0.997 |
| C9 | 1 | 1 | 0.997 |
| C3xD7 | 1 | 1 | 0.940 |
| C23 | 1 | 1 | 1.000 |
| C27 | 1 | 1 | 1.000 |
| C36 | 1 | 1 | 1.000 |
| D20 | 1 | 1 | 1.000 |
| D25 | 1 | 1 | 0.998 |
| C2xC3 | 1 | 1 | 0.980 |
| C7 | 1 | 1 | 0.996 |
| | | | |

| Group | True | Pred | Score |
|---|---|---|---|
| C29 | 1 | 1 | 1.000 |
| S5 | 0 | 0 | 0.409 |
| D2xA5 | 0 | 1 | 0.894 |
| D2xS7 | 0 | 0 | 0.004 |
| C2xA6 | 0 | 0 | 0.222 |
| A6 | 0 | 0 | 0.334 |
| PSL2_5 | 0 | 1 | 0.987 |
| PSL2_7 | 0 | 1 | 0.843 |
| PSL2_11 | 0 | 0 | 0.133 |
| PSL2_13 | 0 | 0 | 0.062 |
| PSL2_17 | 0 | 0 | 0.030 |
| PSL2_19 | 0 | 0 | 0.023 |
| PSL2_23 | 0 | 0 | 0.017 |
| | | | |

The Score column reports the predicted probability assigned to the predicted class and therefore provides a measure of the model's confidence in each prediction. The model correctly classified most test groups, including five of the seven previously unseen $\mathrm{PSL}(2, \mathrm{q})$ groups.

**Table 14.** Confusion matrix and BA of the selected GNN architecture (2-8-8) on the test set.

| Architecture | Solvable | | Non- solvable | | BA |
|---|---|---|---|---|---|
| | TP | FN | TN | FP | |
| (2-8-8) | 28 | 0 | 9 | 3 | 0.875 |

The selected architecture (2-8-8) achieved a BA of 0.875 on the test set. These results demonstrate that the proposed framework can effectively distinguish between solvable and non-solvable groups while exhibiting strong generalization to previously unseen group families.

# 5 Discussion

The experimental results demonstrate that GNNs can successfully predict several fundamental algebraic properties of finite groups directly from their Cayley graph representations. Across the three classification tasks—abelianity, nilpotency, and solvability—the proposed framework consistently achieved balanced accuracies substantially above random guessing, indicating that these algebraic properties leave detectable structural signatures in the corresponding Cayley graphs. The results further demonstrate that a common graph-learning framework can be applied across multiple algebraic properties while varying only the target labeling function.

Among the three classification tasks, abelianity proved to be the easiest to learn, with the best-performing model achieving a perfect test BA of 1.000. The corresponding models for nilpotency and solvability achieved test balanced accuracies of 0.856 and 0.875, respectively. Different GNN architectures proved optimal for different algebraic properties, suggesting that the structural complexity required to learn a property depends on the underlying algebraic structure rather than on a single universally optimal architecture.

For the abelianity and nilpotency tasks, the models correctly classified all previously unseen $\mathrm{PSL}(2,\mathrm{q})$ groups that were excluded entirely from training and validation. For the solvability task, most of the held-out $\mathrm{PSL}(2,\mathrm{q})$ groups were classified correctly. These results suggest that the learned graph representations capture structural characteristics associated with the target properties rather than relying solely on family-specific patterns.

To address the substantial class imbalance in the abelianity and solvability tasks, class-weighted cross-entropy was used during training. Together with BA as the evaluation metric, this provided a more appropriate evaluation framework for the imbalanced classification tasks.

Overall, the results demonstrate that Cayley graph representations encode substantial information about the underlying algebraic properties of finite groups and that GNNs can effectively exploit this information. The present study should be viewed as a proof of concept demonstrating that a unified graph-learning framework can successfully learn multiple algebraic properties directly from Cayley graph representations without requiring property-specific feature engineering. These findings support graph representation learning as a promising methodology for computational group theory and contribute to the growing intersection of group theory, graph theory, and geometric deep learning.

Despite these encouraging results, several limitations remain. First, although the benchmark includes 176 finite groups drawn from multiple classical families and direct-product constructions, it represents only a small subset of all finite groups. Second, the experiments rely on Cayley graphs constructed from specific generating sets, and the effect of alternative generating sets remains to be investigated. Third, although the held-out $\mathrm{PSL}(2, \mathrm{q})$ family provides evidence of generalization beyond the training distribution, additional unseen group families should be considered to obtain a more comprehensive assessment of the framework's generalization capabilities. Finally, future work will investigate additional algebraic properties, larger benchmarks, richer graph representations, and alternative GNN architectures to further improve predictive performance and broaden the applicability of the proposed methodology.

# 6 Conclusion

In this work, we presented a property-independent GNN framework for predicting algebraic properties of finite groups directly from their Cayley graph representations. Using a common graph construction procedure, feature representation, GNN architecture, and training pipeline, the framework was successfully applied to the prediction of abelianity, nilpotency, and solvability. The best-performing models achieved test balanced accuracies of 1.000, 0.856, and 0.875, respectively, demonstrating that GNNs can effectively learn multiple algebraic properties from Cayley graph representations.

Overall, this work provides a proof of concept that graph representation learning offers a unified approach to predicting algebraic properties of finite groups. Future work will extend the framework to additional algebraic properties and group families, investigate alternative Cayley graph constructions and generating sets, and evaluate richer graph representations and more expressive GNN architectures on larger benchmark datasets.